\def\year{2021}\relax
\documentclass[letterpaper]{article} %
\usepackage{aaai21}  %
\usepackage{times}  %
\usepackage{helvet} %
\usepackage{courier}  %
\usepackage[hyphens]{url}  %
\usepackage{graphicx} %
\usepackage{natbib}  %
\usepackage{caption} %
\usepackage{todonotes}

\usepackage{wrapfig}

\usepackage[utf8]{inputenc} %
\DeclareUnicodeCharacter{2212}{-}

\usepackage{microtype}
\usepackage{graphicx}
\usepackage{booktabs} %
\usepackage{multirow}
\usepackage{siunitx}
\usepackage[frozencache,cachedir=.]{minted}

\usepackage{fancyhdr}
\usepackage{lastpage}

\usepackage[hyphens]{url}
\usepackage[]{hyperref}
\hypersetup{breaklinks=true}

\usepackage{pgfplots}
\pgfplotsset{compat=newest}
\usepgfplotslibrary{groupplots}
\usepgfplotslibrary{dateplot}

\usepackage{caption}
\usepackage{subcaption}

\usepackage{appendix}

\usepackage{tikz}
\usetikzlibrary{decorations.text,calc,shapes,arrows,arrows.meta, positioning,shapes.misc,decorations.markings,decorations.markings,decorations.pathreplacing,matrix,spy}

\usepackage{amsmath,bm}
\usepackage{amssymb}

\usepackage[autostyle, english=american]{csquotes}
\MakeOuterQuote{"}

\everypar{\looseness=-1} %
\usepackage{adjustbox}

\usepackage{hyperref}

\begin{document}

\title{Research Reproducibility as a Survival Analysis}

\author{%
  Edward Raff \textsuperscript{\rm 1,2}\\
  }
  
\affiliations {
    \texttt{raff\_edward@bah.com}, \texttt{raff.edward@umbc.edu} \\
    \textsuperscript{\rm 1} Booz Allen Hamilton,
    \textsuperscript{\rm 2} University of Maryland, Baltimore County
}

\maketitle

\begin{abstract}
There has been increasing concern within the machine learning community that we are in a reproducibility crisis. As many have begun to work on this problem, all work we are aware of treat the issue of reproducibility as an intrinsic binary property: a paper \textit{is} or \textit{is not} reproducible. Instead, we consider modeling the reproducibility of a paper as a survival analysis problem. We argue that this perspective represents a more accurate model of the underlying meta-science question of reproducible research, and we show how a survival analysis allows us to draw new insights that better explain prior longitudinal data. 
The data and code can be found at \url{https://github.com/EdwardRaff/Research-Reproducibility-Survival-Analysis}
\end{abstract}

\section{Introduction}
\label{sec:intro}

There is current concern that we are in the midst of a reproducibility crisis within the fields of Artificial Intelligence (AI) and Machine Learning (ML) \cite{Hutson725}. Rightfully, the AI/ML community has done research to understand and mitigate this issue. Most recently, \citet{Raff2019_quantify_repro} performed a longitudinal study attempting to independently reproduce 255 papers, and they provided features and public data to begin answering questions about the factors of reproducibility in a quantifiable manner. Their work, like all others of which we are aware, evaluated reproducibility using a binary measure. Instead, we argue for and demonstrate the value of using a survival analysis. In this case, we model the hazard function $\lambda(t | \boldsymbol{x})$, which predicts the likelihood of a an event (i.e., reproduction) occurring at time $t$, given features about the paper $\boldsymbol{x}$. 
In survival analysis, we want to model the likely time until the occurrence of an event. This kind of analysis is common in medicine where we want to understand what factors will prolong a life (i.e., increase survival), and what factors may lead to an early death. 

In our case, the event we seek to model is the successful reproduction of a paper's claims by a reproducer that is independent of the original paper's authors. Compared to the normal terminology used, we desire factors that \textit{decrease survival} meaning a paper takes \textit{less time} to reproduce. In the converse situation, a patient that "lives forever" would be equivalent to a paper that cannot be reproduced with any amount of effort. We will refer to this as \textit{reproduction time} in order to reduce our use of standard terminology that has the opposite connotation of what we desire. Our goal, as a community, is to reduce the reproduction time toward zero and to understand what factors increase or decrease reproduction time. 

We will start with a review of related work on reproducibility, specifically in machine learning, in \autoref{sec:related_work}. Next, we will detail how we developed an extended dataset with paper survival times in \autoref{sec:data}. We will use models with the Cox proportional hazards assumption to perform our survival analysis, which will begin with a linear model in \autoref{sec:linear_cox}. The linear model will afford us easier interpretation and statistical tools to verify that the Cox proportional hazards model is reasonable for our data. In \autoref{sec:non_linear} we will train a non-linear Cox model that is a better fit for our data in order to perform a more thorough analysis of our data under the Cox model. Specifically, we show in detail how the Cox model allows us to better explain observations originally noted in \cite{Raff2019_quantify_repro}. 
This allows us to measure a meaningful effect size for each feature's impact on reproducibility, and better study their relative importance's. We stress that the data used in this study does not mean these results are definitive, but useful as a new means to think about and study reproducibility. 

\section{Related Work} \label{sec:related_work}

The meta-science question of reproducibility --- scientifically studying the research process itself --- is necessary to improve how research is done in a grounded and objective manner \cite{Ioannidis2018}. Significant rates of non-reproduction have been reported in many fields, including a 6\% rate in clinical trials for drug discovery \cite{Prinz2011}, making this an issue of greater concern in the past few decades. Yet, as far as we are aware, all prior works in machine learning and other disciplines treat reproducibility as a binary problem \cite{Gundersen2018,Glenn2015,Ioannidis2017,Wicherts2006}. Even works that analyze the difference in effect size between publication and reproduction still view the issue result in a binary manner \cite{aac4716}.  

Some have proposed varying protocols and processes  that authors may follow to increase the reproducibility of their work \cite{Barba2019,Gebru2018}. While valuable, we lack quantification of their effectiveness. In fact, little has been done to empirically study many of the factors related to reproducible research, with most work being based on a subjective analysis. \citet{Olorisade2018} performed a small scale study over 6 papers in the specific sub-domain of text mining. \citet{pmlr-v97-bouthillier19a} showed how replication (e.g., with docker containers), can lead to issues if the initial experiments 
use fixed seeds, which has been a focus of other work \cite{Forde2018}. The largest empirical study was done by \citet{Raff2019_quantify_repro}, which documented features while attempting to reproduce 255 papers. This study is what we build upon for this work.  

\citet{Sculley2018} have noted a need for greater rigor in the design and analysis of new algorithms. They note that the current focus on empirical improvements and structural incentives may delay or slow true progress, despite appearing to improve monotonically on benchmark datasets. This was highlighted in recent work by \citet{Dacrema2019} who attempted to reproduce 18 papers from Neural Recommendation algorithms. Their study found that only 7 could be reproduced with "reasonable effort." Also concerning is that 6 of these 7 could be outperformed by better tuning baseline comparison algorithms. These issues regarding the nature of progress, and what is actually learned, are of extreme importance. We will discuss these issues as they relate to our work and what is implied about them from our results. However, the data from \cite{Raff2019_quantify_repro} that we use does not quantify such "delayed progress," but only the reproduction of what is stated in the paper. Thus, in our analysis, we would not necessarily be able to discern the issues with the 6 papers with insufficient baselines by \cite{Dacrema2019}. 

We note that the notion of time impacting reproduction by the \textit{original authors }or using \textit{original code} has been previously noted \cite{Mesnard2017,10.1371/journal.pone.0038234} and often termed "technical debt" \cite{10.5555/2969442.2969519}. While related, this is a fundamentally different concern. Our study is over independently attempted implementation, meaning technical debt of an existing code base can not exist. Further, these prior works still treat replication as a binary question despite noting the impact of time on difficulty. Our distinction is using the time to implement itself as a method of quantifying the degree or difficulty of replication, which provides a meaningful effect size to better study replication. 

\section{Study Data} \label{sec:data}

\begin{figure}[!htbp]
\centering
\adjustbox{max width=\columnwidth}{%
    \input{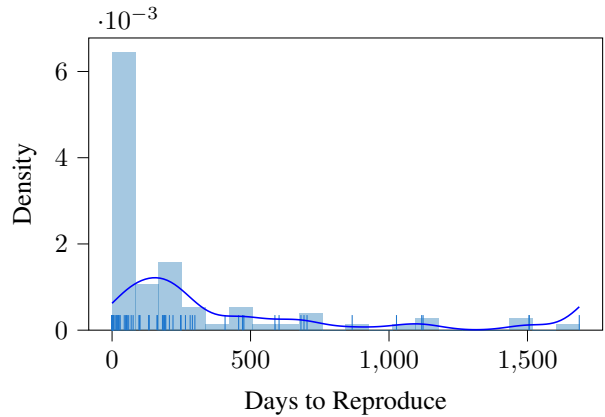}
}
\caption{Histogram of the time taken to reproduce.
The dark blue line shows a Kernel Density Estimate of the density, and the dashes on the x-axis indicate 
specific values.
} \label{fig:repro_time_stats}
\end{figure}

The original data used by \citet{Raff2019_quantify_repro} was made public but with explicit paper titles removed. We have augmented this data in order to perform this study. Specifically, of the papers that were reproduced, the majority had their implementations made available as a part of the JSAT library \cite{JMLR:v18:16-131}. Using the Github history, we were able to determine end dates for the completion of an algorithm's implementation. In addition the original start dates of each implementation are recorded by Mendeley which were used for the original study. Combined, this gives us start and end dates and survival times for 90 out of the 162 papers that were reproduced. The remaining 44\% of reproduced papers, for which we could not determine any survival time, were unfortunately excluded from the analysis conducted in the remainder of this paper. Summary statistics on the time taken to reproduce these 90 papers are shown in \autoref{fig:repro_time_stats}. While most were reproduced quickly, many papers required months or years to reproduce. 

\citet{Raff2019_quantify_repro} noted that the attempts at implementation were not continuous efforts, and noted many cautions of potential bias in results (e.g., most prominently all attempts are from a single author). Since we are extending their data, all these same biases apply to this work --- with additional potential confounders. Total amount of time to implement could be impacted by life events (jobs, stresses, etc.), lack of appropriate resources, or attempting multiple works simultaneously, which is all information we lack. As such readers must temper any expectation of our results being a definitive statement on the nature of reproduction, but instead treat results as initial evidence and indicators. 
Better data must be collected before stronger statements can be made.
Since the data we have extended took 8 years of effort, we expect larger cohort studies to take several years of effort.
We hope our paper
will encourage such studies to include time spent into the study's design and motivate participation and compliance of cohorts. 

With appropriate caution, our use of Github as a proxy measure of time spent gives us a level of ground truth for the reproduction time for a subset of reproduced papers. We lack any labels for the time spent on papers which failed to be reproduced and successful reproductions outside of Github. Survival analysis allows us to work around some of these issues as a case of \textit{right censored} data. Right censored data would indicate a start time $s$, an amount of time observed $t_o$, but not the successful event (reproduction) $e$. If $t_e$ is the amount of time needed to observe the event $e$ (i.e., true amount of time needed to reproduce), then we have $t_o < t_e$. If we can estimate a minimum time $0 < \hat{t_o} < t_o$, we can still perform a reasonable and meaningful survival analysis. We work under the assumption that more time was likely spent on papers that were not reproduced than ones that were, and so assume that the average time spent on successful papers ($\hat{t_o}$) than the actual amount of time spent $(t_o)$. As such, we assign every non-reproduced paper the average amount of time for the data in \autoref{fig:repro_time_stats}, or 218.8 days of effort, as an approximate guess at the amount of time that would be expended. We note that we found our analysis to be robust to a wide range of alternative constants, such as the median time (53.5 days). While the exact values in the analysis would change, the results were qualitatively the same. A repetition showing qualitative similarity of using the median can be found in \autoref{sec:by_median}. 

A survival model does not provide us the means to circumvent the cases where we lack the observed time $t_o$ for successfully reproduced papers. Due to the reduced number of data points (the 72 papers not implemented in JSAT's Github), we have excluded some of the variables from analysis in this paper. Specifically, the Venue a paper was published in and the Type of Venue. This reduced dataset resulted in significant skew in some of the sub-categories of these fields, making comparisons untenable (e.g., the number of papers in Workshops was reduced to one example).

The above protocol was used to create the data for this study, and will be made available for others to study and perform additional analysis on. Our lack of ground truth "level of effort" for all of the original data is a source of potential bias in our results, and should be used to caution on taking the results as any kind of proclamation or absolute truth. That said, we find the analysis useful and able to elucidate finer details and insights that were not recognizable under the standard binary analysis of reproduction. 

\section{Linear Hazard Analysis} \label{sec:linear_cox}

We start with standard Cox model, where we use a linear set of coefficients $\boldsymbol{\beta} \in \mathbb{R}^d$ to control how much our various features impact the predicted reproduction time,
$\lambda(t | \boldsymbol{x}_i) = \exp\left(\boldsymbol{x}_i^\top \boldsymbol{\beta} \right) \cdot \lambda_0(t)$, where 
$\lambda_0(t)$ is a baseline hazard function. The Cox model imposes a \textit{proportional} assumption on the nature of the hazard. The base hazard rate $\lambda_0(t)$ may by different for every individual, but the proportional assumption means that we expect the impact of altering any covariate  to have the same proportional effect for every instance. e.g., specifying the hyperparamters used in a paper would always reduce the reproduction time by a factor of X\%.

The Cox model provides us a number of benefits to our analysis, provided that it is a reasonable assumption. First, it allows us to estimate $\boldsymbol{\beta}$ without knowing, or even modeling, the base hazard function $\lambda_0(t)$. Second, though not unique to the Cox model, it supports \textit{right censored} data. If an instance is right censored, we have waited for the event to occur for some unit of time $t$, but have not yet seen the event occur at a later point in time $t_e$. This allows us to model all of the failed reproduction attempts as papers which may be reproducible, but for which we have not yet put in sufficient effort to reproduce the paper. This allows us to use our estimated effort spent on non-reproduced papers without causing significant harm to the underlying model.

\subsection{Cox Proportional Hazard Validation}

\begin{table}[!htbp]
\caption{p-values for each features' importance.
}
\label{tbl:p_vals}
\adjustbox{max width=\columnwidth}{%
\begin{tabular}{@{}lrrr@{}}
\toprule
\multicolumn{1}{c}{Feature} & \multicolumn{1}{c}{Independent} & \multicolumn{1}{c}{Logistic} & \multicolumn{1}{c}{Cox}   \\ \midrule
Year Published              & 0.964                           & 0.613                        & 0.92                      \\
Year Attempted              & 0.674                           & 0.883                        & 0.45                      \\
Has Appendix                & 0.330                           & 0.201                        & \textbf{0.07}             \\
Uses Exemplar Toy Problem   & 0.720                           & 0.858                        & 0.20                      \\
Looks Intimidating          & 0.829                           & \textbf{0.035}               & 0.20                      \\
\textit{Exact Compute Used }         & 0.257                           & 1.000                        & 0.39                      \\
\textit{Data Available}              & \textbf{\textless{}0.005}       & 0.644                        & 0.81                      \\
Code Available              & 0.213                           & 0.136                        & 0.18                      \\
Number of Authors           & 0.497                           & 0.542                        & 0.68                      \\
Pages                       & 0.364                           & 0.702                        & 0.82                      \\
Num References              & 0.740                           & 0.421                        & 0.54                      \\
Number of Equations         & \textbf{0.004}                  & 0.478                        & 0.74                      \\
Number of Proofs            & 0.130                           & 0.629                        & 0.47                      \\
Number of Tables            & \textbf{0.010}                  & 0.618                        & 0.86                      \\
Number of Graphs/Plots      & 0.139                           & 0.325                        & 0.51                      \\
Number of Other Figures     & 0.217                           & 0.809                        & 0.98                      \\
Conceptualization Figures   & 0.365                           & 0.349                        & 0.13                      \\
Hyperparameters Specified   & \textbf{\textless{}0.005}       & 1.000                        & 0.49                      \\
Algorithm Difficulty        & \textbf{\textless{}0.005}       & 1.000                        & \textbf{0.03}             \\
Paper Readability           & \textbf{\textless{}0.005}       & 1.000                        & \textbf{\textless{}0.005} \\
Pseudo Code                 & \textbf{\textless{}0.005}       & 1.000                        & \textbf{\textless{}0.005} \\
Compute Needed              & \textbf{\textless{}0.005}       & 1.000                        & 1.000                     \\
Rigor vs Empirical          & \textbf{\textless{}0.005}       & 1.000                        & \textbf{0.02}             \\ \bottomrule
\end{tabular}
}
\end{table}

The question we must first answer: is the Cox proportional model assumption reasonable for our data? First, as a baseline, we have the results of the non-parametric tests performed by \cite{Raff2019_quantify_repro}. Next, we train a linear logistic regression model and a linear Cox proportional hazards model on our reduced subset, and we compute and compare 
which features were found as significant.
Due to the impact of outliers in our data, we found it necessary to use a robust estimate of the Cox linear model \cite{Lin1989}. Last, we perform simple cross-validation to show reasonable predictive performance. 

We highlight here that our goal is not to use these results as a new determination for what factors are important, but as a method of comparing the relative appropriateness and agreement between a logistic and Cox model with that of the original analysis of \cite{Raff2019_quantify_repro}. The two new models are trained on less data. In addition, the logistic and Cox models required a regularization penalty to be added to the model due to the data being linearly separable, which prevented convergence. This means the p-values presented are not technically accurate. For the logistic and Cox models, the test is whether the parameter in question has a non-zero slope, which is impacted by the values of all other features.

\autoref{tbl:p_vals} shows the resulting p-values. The logistic model finds only one significant relationship, which is not corroborated by the data or previous analysis. In addition, it marks several features as unimportant ($p$=1.0) that were previously found to be highly significant. This shows that a classification model of reproducibility is broadly not appropriate for modeling reproducibility. In general, there is considerable agreement between the Cox model and the original non-parametric testing of individual factors. These results in isolation give us an increased  confidence that the Cox proportional hazards assumption is reasonable, as it has reasonable correspondence with the original analysis. We note that if one instead performs the Independent tests on the same sub-set of data (due to aforementioned 72 papers removed), the results are robust. Only the two features \textit{italicized} change, with \textit{Exact Compute} becoming significant and \textit{Code Available} becoming non-significant.

If the proportional hazard assumption is correct, we should observe that the hazard ratio over time for any two points $i$ and $j$ should remain a constant for all time $t$.
This can be tested using the Kaplan–Meier (KM) \cite{Kaplan1958} and log-rank \cite{PMID:5910392} tests.

\begin{table}[!htbp]
\caption{Statistical Test of Cox proportional hazard assumption for each feature, shown for the two cases that rejected the null hypothesis. }
\label{tbl:cox_assumption_test_failed}
\centering
\adjustbox{max width=\columnwidth,max height=0.75\textheight}{%
\begin{tabular}{@{}llcS[table-format = <1.3]@{}}
\toprule
\multicolumn{1}{c}{Feature}                           & \multicolumn{1}{c}{Transform} & Statistic & {p-value}          \\ \midrule
\multirow{2}{*}{\begin{tabular}[c]{@{}l@{}}Normalized Number\\ of Equations\end{tabular}}       & km                            & 4.30      & 0.04             \\
                                                      & rank                          & 8.03      & \textless{}0.005 \\
                                                      \cmidrule(l){2-4} 
\multirow{2}{*}{Year Attempted}                       & km                            & 5.78      & 0.02             \\
                                                      & rank                          & 8.76      & \textless{}0.005 \\ \bottomrule
\end{tabular}
}
\end{table}

We tested this assumption for each feature, and found only two that rejected the null hypothesis. This was the Normalized Number of Equations and Year Attempted features, as shown in \autoref{tbl:cox_assumption_test_failed}.  We note that in isolation only two features failing out of 34 coefficients in the model is encouraging, and it corresponds with having $34\cdot0.05=1.7$ false positives at a significance level of $\alpha=0.05$. 
We also investigate these two features further by evaluating their residual fits under the linear model
(see appendix \autoref{fig:cox_residual_fits}). 
In both cases, we can see that the Cox model is actually a decent fit, with the residual errors concentrated about zero. The failure appears to be due to the linear nature of the current Cox model, which we address in \autoref{sec:non_linear}.

The agreement between the Cox model on significant factors, as well as the Cox model passing goodness of fit tests, are all indicators of the appropriateness of this approach. As such, it is not surprising that we would obtain reasonable predictive performance with the Cox model. Specifically, we measure the Concordance \cite{HarrellJr.1996} for the linear model,
defined as $\frac{2 C + T}{2 (C + D + T)}$.
Here $C$ refers to the number of pairs where there model correctly predicts the survival order (i.e., which paper was reproduced first is ordered first), $D$ the number of pairs where the model predicted the wrong order, and $T$ the number of ties made by the model. 
Under Concordance, a random model would receive a score of 0.5. Using 10-fold cross validation, the linear Cox model presented here obtains 0.73. 

\subsection{Analysis of Linear Cox Model}

\begin{table}[!htb]
\caption{Coefficients of the linear Cox hazard model. }
\label{tbl:cox_linear_coefs}
\adjustbox{max width=\columnwidth}{%
\begin{tabular}{@{}lS[table-format=1.2]S[table-format=1.2]S[table-format=1.2]@{}}
\toprule
Feature                              & {$\beta_i$}  & {$\exp(\beta_i)$} & {stnd. err} \\ \midrule
Year                                 & -0.00 & 1.00      & 0.03     \\
Year Attempted                       & -0.10 & 0.91      & 0.13     \\
Has Appendix                         & -0.57 & 0.57      & 0.31     \\
Uses Exemplar Toy Problem            & -0.45 & 0.64      & 0.35     \\
Exact Compute Used                   & 0.30  & 1.35      & 0.35     \\
Looks Intimidating                   & 0.47  & 1.61      & 0.37     \\
Data Available                       & -0.12 & 0.89      & 0.50     \\
Author Code Available                & 0.42  & 1.52      & 0.31     \\
Number of Authors                    & 0.03  & 1.04      & 0.09     \\
Pages                                & 0.01  & 1.01      & 0.02     \\
Normalized Num References            & 0.09  & 1.10      & 0.15     \\
Normalized Number of Equations       & 0.02  & 1.02      & 0.07     \\
Normalized Number of Proofs          & 0.42  & 1.52      & 0.58     \\
Normalized Number of Tables          & -0.13 & 0.88      & 0.73     \\
Normalized Number of Graphs/Plots    & -0.08 & 0.92      & 0.12     \\
Normalized Number of Other Figures   & 0.02  & 1.02      & 1.06     \\
Normalized Conceptualization Figures & 1.28  & 3.59      & 0.84     \\
Hyperparameters Specified\_No        & -0.13 & 0.87      & 0.25     \\
Hyperparameters Specified\_Partial   & 0.53  & 1.70      & 0.76     \\
Hyperparameters Specified\_Yes       & 0.03  & 1.03      & 0.20     \\
Paper Readability\_Excellent         & 1.49  & 4.46      & 0.37     \\
Paper Readability\_Good              & 0.99  & 2.70      & 0.22     \\
Paper Readability\_Ok                & -0.17 & 0.85      & 0.31     \\
Paper Readability\_Low               & -1.29 & 0.27      & 0.24     \\
Algo Difficulty\_High                & -0.02 & 0.98      & 0.23     \\
Algo Difficulty\_Medium              & -0.42 & 0.66      & 0.20     \\
Algo Difficulty\_Low                 & 0.44  & 1.56      & 0.23     \\
Pseudo Code\_Code-Like               & -0.03 & 0.97      & 0.45     \\
Pseudo Code\_Yes                     & -0.02 & 0.98      & 0.20     \\
Pseudo Code\_Step-Code               & -0.65 & 0.52      & 0.35     \\
Pseudo Code\_No                      & 0.68  & 1.97      & 0.21     \\
Rigor vs Empirical\_Balance          & 0.45  & 1.57      & 0.19     \\
Rigor vs Empirical\_Empirical        & -0.15 & 0.86      & 0.35     \\
Rigor vs Empirical\_Theory           & -0.32 & 0.73      & 0.32     \\ \bottomrule
\end{tabular}
}
\end{table}

Given the above results, it is clear that the Cox proportional hazards model is an appropriate method to model the joint set of features and analyze how they impact reproducibility time. The assumptions of the Cox model appear to hold for our data, and the p-values indicate that the model has correspondence with the original analysis as well as  superior fit compared to a simpler logistic approach that treats the problem as a binary classification problem. 

While we will show later that a non-linear Cox approach better models the data, we find it instructive to first examine the insights and results of the linear approach as it is easier to interpret. In \autoref{tbl:cox_linear_coefs}, we present the coefficients $\beta_i$ for each variable, its exponentiated value $\exp(\beta_i)$, and the estimated standard error for each coefficient. 
The value $\exp(\beta_i)$ is of particular relevance, as in the linear Cox model it can be interpreted as the proportional impact on the paper's reproduction time per unit increase in the feature. For example, this table indicates the Year a paper was published has no impact with a value $\exp(\beta_\text{Year}) = 1$. 

In \citet{Raff2019_quantify_repro}, it was found that a paper's readability was a significant relationship. While the obvious interpretation of these results was that more readable papers are more reproducible, the binary measure of reproducibility used gives us no means to quantify the nature of this relationship. We can now quantify this nature under the framework of survival analysis. Papers that had "Excellent" readability reduced the reproduction time by 4.46$\times$ and "Good" papers by 2.70$\times$. Those that where "Ok" begin to take longer, increasing the time to reproduce by $0.85^{-1}\approx1.18\times$ and "Low" readability by $0.27^{-1}\approx 3.70\times$.

In examining the coefficients presented here, we highlight the importance of considering the standard error. For example, the second most impactful feature appears to be the number of Conceptualization Figures per page, but the standard error is nearly the same magnitude as the coefficient itself. This indicates that the relationship may not be as strong as first indicated by the linear Cox model.

\begin{figure*}[!htb]
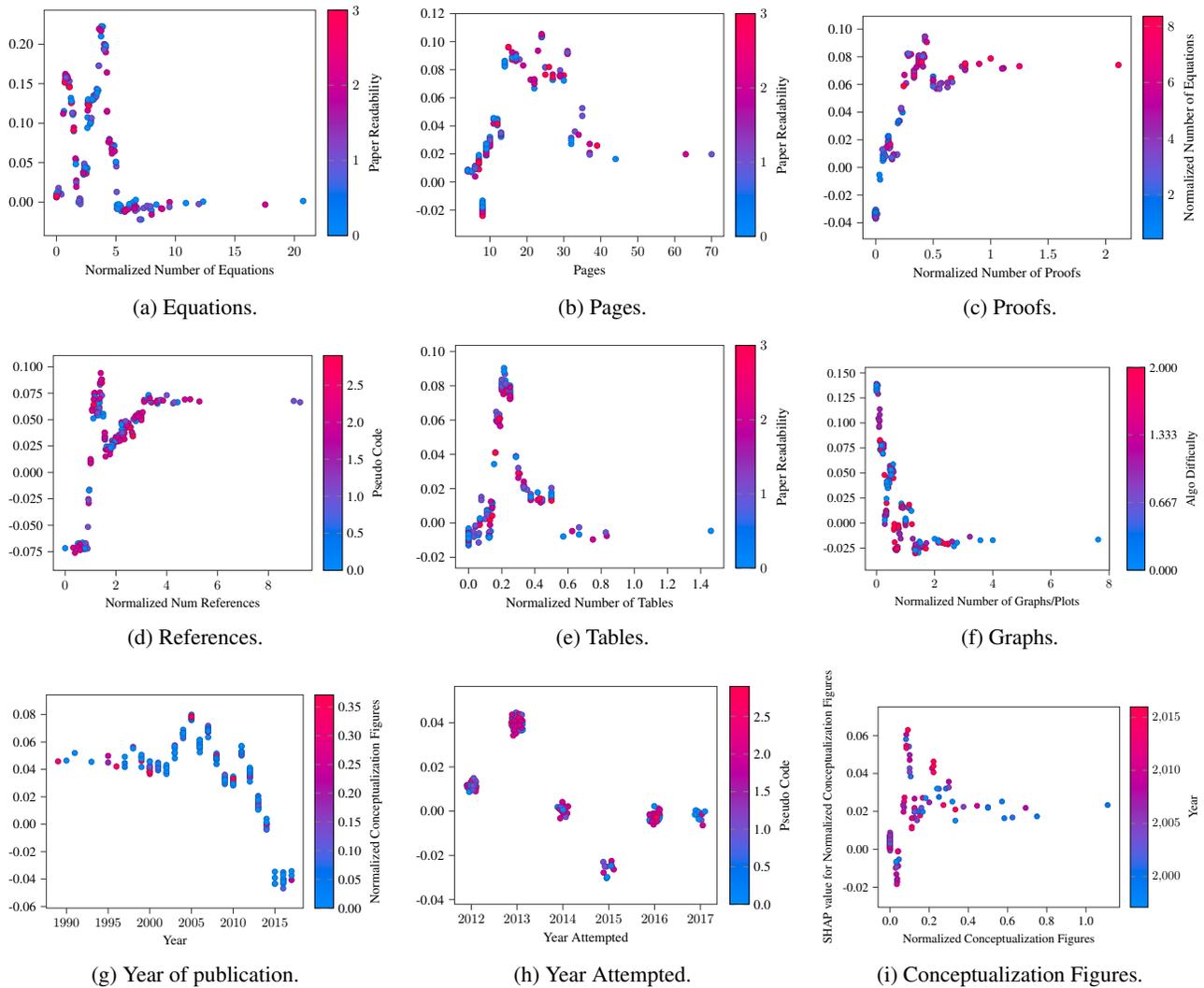

\centering
\setlength{\lineskip}{\medskipamount}
\subcaptionbox{Equations.\label{fig:xgboost_equations}}
    {
        \adjustbox{max width=0.311\textwidth}{%
	    \input{figs/shap/xgboost_shape_decor_Normalized_Number_of_Equations.tex}
	    }
    }
\subcaptionbox{Pages.\label{fig:xgboost_pages}}
    {
        \adjustbox{max width=0.311\textwidth}{%
	    \input{figs/shap/xgboost_shape_decor_Pages.tex}
	    }
    }
\subcaptionbox{Proofs.\label{fig:xgboost_proofs}}
    {
        \adjustbox{max width=0.311\textwidth}{%
	    \input{figs/shap/xgboost_shape_decor_Normalized_Number_of_Proofs.tex}
	    }
    }
\subcaptionbox{References.\label{fig:xgboost_references}}
    {
        \adjustbox{max width=0.311\textwidth}{%
	    \input{figs/shap/xgboost_shape_decor_Normalized_Num_References.tex}
	    }
    }
\subcaptionbox{Tables.\label{fig:xgboost_tables}}
    {
        \adjustbox{max width=0.311\textwidth}{%
	    \input{figs/shap/xgboost_shape_decor_Normalized_Number_of_Tables.tex}
	    }
    }
\subcaptionbox{Graphs.\label{fig:xgboost_graphs}}
    {
        \adjustbox{max width=0.311\textwidth}{%
	    \input{figs/shap/xgboost_shape_decor_Normalized_Number_of_Graphs_Plots.tex}
	    }
    }
\subcaptionbox{Year of publication.\label{fig:xgboost_year}}
    {
        \adjustbox{max width=0.311\textwidth}{%
	    \input{figs/shap/xgboost_shape_decor_Year.tex}
	    }
    }
\subcaptionbox{Year Attempted.\label{fig:xgboost_year_attempted}}
    {
        \adjustbox{max width=0.311\textwidth}{%
	    \input{figs/shap/xgboost_shape_decor_Year_Attempted.tex}
	    }
    }
\subcaptionbox{Conceptualization Figures.\label{fig:xgboost_conceptualization}}
    {
        \adjustbox{max width=0.311\textwidth}{%
	    \input{figs/shap/xgboost_shape_decor_Normalized_Conceptualization_Figures.tex}
	    }
    }

\caption{SHAP individual features (change in log-hazard ratio) for several numeric features on the y-axis with feature values on the x-axis. Each figure has a color set based on the value of a second feature, indicated on the right. The second feature is selected by having the highest SHAP interaction. 
} \label{fig:xgboost_cox_individuals}
\end{figure*}

\section{Non-Linear Cox and Interpretation} \label{sec:non_linear}

We have now used the linear Cox model to evaluate the appropriateness of a Cox model, where we can more readily compare the Cox Proportional Hazard assumptions with the independent statistical testing of prior work, as well as a logistic alternative. Now, we use a non-linear tree-based model of the Cox assumption and focus our evaluation on the results, and their meaning, with this non-linear model. We do this because the data exhibits non-linear behavior and interactions that made it difficult to fit the linear Cox model. This prevents us from using standard statistical tests on the coefficients with full confidence, and we know that outliers and non-linear behavior plays a non-trivial impact on the results of the Cox regression,  
$\lambda(t | \boldsymbol{x}_i) = \exp\left(f(\boldsymbol{x}_i) \right) \cdot \lambda_0(t)$.

In particular, the XGBoost
library implements a Cox log-likelihood based splitting criteria\cite{Bou-Hamad2011,LeBlanc1992}, allowing us to use a boosted decision tree to model the $\exp(\boldsymbol{x}_i^T \boldsymbol{\beta})$ term of the standard linear Cox model. Optuna \cite{Akiba:2019:ONH:3292500.3330701} was used to tune the parameters of the model resulting in a 10-fold cross-validated Concordance score of 0.80, which is a significant improvement over the linear Cox model indicating a better fit. For this model, we encoded X features as ordinal values that correspond with the nature of what they measure since XGBoost does not support multinomial features. This included the Paper Readability (\textit{Low}=0, \textit{Ok}=1, \textit{Good}=2, \textit{Excellent}=3), Algorithm Difficulty (\textit{Low}=0,\textit{Medium}=1, \textit{High}=2), and Pseudo Code (\textit{None}=0, \textit{Step-Code}=1, \textit{Yes}=2, \textit{Code-like}=3) features. All others were encoded in a one-hot fashion.
\begin{figure}[!htb]
    \centering
    \includegraphics[width=\columnwidth]{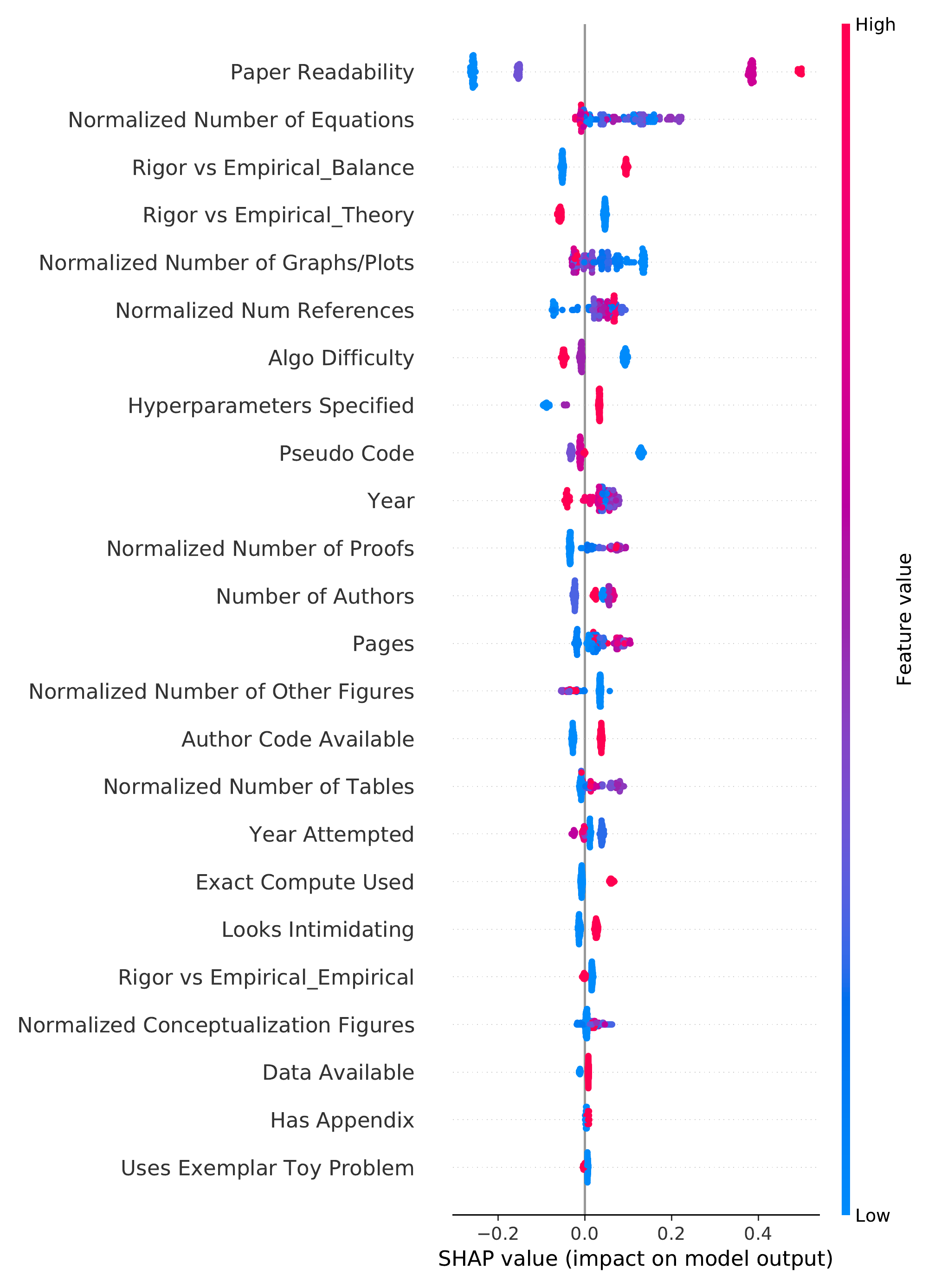}
    \caption{
    SHAP results for each feature. 
    The change in log hazard ratio (x-axis) caused by each feature where 
    the color coding indicates the values from high (red) to low (blue).}
    \label{fig:xgboost_shap_decor}
\end{figure}

With a boosted survival tree as our model, we can use SHapley Additive exPlanations (SHAP) \cite{NIPS2017_7062} as a method of understanding our results in greater detail and nuance than the linear Cox model allows. The SHAP value indicates the change in the model's output contributed by a specific feature value. This is measured for each feature of each datum, giving us a micro-level of interpretability, while we can obtain a macro-level by looking at the average or distribution of all SHAP values. The SHAP value unifies several prior approaches to interpreting the impact of a feature $\boldsymbol{x}$'s values on the output  \cite{Ribeiro:2016:WIT:2939672.2939778,Strumbelj:2014:EPM:2687513.2687579,Shrikumar2017,Datta2016,Bach2015,Lipovetsky2001}. In addition, \citet{Lundberg2018} showed how to exactly compute a tree's SHAP score in $O(L D^2)$ time, where $L$ is the maximum number of leaf nodes and $D$ the maximum depth. The Tree SHAP is of particular value because it allows us to disentangle the impact of feature interactions within the tree, giving us a feature importance that makes it easier to understand on an an individual basis with less complication.

A simple plot of all the SHAP values, sorted by their importance, is presented in \autoref{fig:xgboost_shap_decor}. Red indicates higher per-feature values and blue lower. The x-axis SHAP values are changes in the log hazard ratio. Positive SHAP values correspond to features decreasing the reproduction time, while negative values indicate the feature correlates with an increase the time to reproduce. For example, a SHAP score of 0.4 would be a $\exp(0.4) \approx 1.49 = 49\%$ reduction in the time to reproduce. At a high level, we can immediately see that a number of the features identified as important in \citet{Raff2019_quantify_repro} are also important according the Cox model, with the same overall behaviors. Paper Readability, Number of Equations, Rigor vs Empirical, Algorithm Difficulty, Hyperpramaters Specification, and Pseudo-Code are all important. Other features such as the Use of an Exemplar Toy Problem, Having an Appendix, Data Made Available, or Looking Intimidating have no significant impact. 

We also see what may at first appear to be contradictions with prior analysis. The number of Graphs/Plots was originally determined to have a non-significant relationship. In our model having fewer Graphs/Plots appears to be one of the most significant factors. We explain this by again noting the distinction of our model: the Cox modeling presented above indicates the impact on \textit{how long} it will take us to reproduce the result, whereas the result by \citet{Raff2019_quantify_repro} indicates only if the feature has a significant impact on the \textit{success} of reproduction. As such, we argue there is no contradiction in the results but only deeper insight into their nature. In the case of Graph/Plot, we say that the number of graphs or plots does not impact whether a paper can be reproduced; however, if it is reproducible then having too many graphs and plots increases the time it takes to reproduce. 

Due to space limitations, we can not discuss all of the behaviors that are observable from this data. For this reason, we will attempt to avoid discussing results that reaffirm previous conclusions. We will discuss features in this context if the new Cox analysis provides a particular new insight or nuance into the nature of the relationship between said feature and reproducibility. 

We note that most categorical features have a relatively simple analysis, and the results previously found to be significant also have the most impact on reproduction time. For example, one can clearly see that \textit{Excellent} paper readability decreases reproduction time while \textit{Low} readability increases it. Results are also consistent with the sign of correlation previously found, e.g., \textit{Theory} papers take more time to reproduce than \textit{Empirical} ones. However, new insight is derived from noting that \textit{Balanced} papers further reduce the reproduction time.

Of these categorical variables, we draw special attention to three results. First, the nature of including pseudo-code is surprising. Before, it was noted that \textit{Step-Code} negatively correlated with reproducibility, but papers with \textit{No} and \textit{Code-Like} pseudo-code were more reproducible. Under the Cox model, papers with \textit{No} pseudo-code take significantly less time to reproduce.  Step-Code is found to have a negative impact, while simply having pseudo-code or \textit{Code-Like} pseudo-code have almost no impact on the hazard ratio. 

We also draw attention to the fact that specifying the Exact Compute Used and the authors making their code available lead to reductions in compute time. As part of the original study, all papers with Code Available were excluded if the code was used prior to reproduction. We suspect that the reduction in reproduction time is a psychological one: that more effort is expended with higher confidence that the results can be replicated when the authors make their code available. A similar hypothesis might be drawn by specifying the Exact Compute Used leading to a reduced reproduction time. Alternatively, it may be the case that the exact equipment provides useful information for the reproducer to judge the meaning behind discrepancies in performance and to adjust accordingly. For example, if the nature of a proposed speedup is dependent on reducing communication overhead between CPU threads, and the reproducer has a system with slower per-core clock speeds than the paper, it could explain getting reduced results --- and that final replication may need to be tested on different equipment.

\textbf{Detailed Analysis of Numerical Features:}
It was originally noted that the features determined to be significantly related to reproducibility were also the most subjective, which reduces the utility of the results. Under the boosted Cox model, we can more easily and readily derive insights from the more objective features. In \autoref{fig:xgboost_cox_individuals}, we provide individual plots of all the numeric features. In each case, the SHAP value is plotted on the y-axis and a color is set based on the value of a second feature. The second feature is selected based on the SHAP interaction scores, where the feature with the highest mean interaction score is used to color the plot. This provides additional insights for some cases. However, the degree of non-linear interaction between features is relatively small so the coloring is not always useful.

With respect to the length of a paper (\autoref{fig:xgboost_pages}), we see that there is a strong positive correlation with paper length, with most Journal papers having reduced reproduction time. 
This relationship ends with  excessively long papers ($\geq$35 pages) that return the the base hazard rate.
We believe this relationship stems from the observation that page limits may discourage more implementation details \cite{Raff2019_quantify_repro}. 
Under the Cox model, it becomes clearer that larger page limits in conferences, or more emphasis on Journal publications, is of value in reducing the time it takes to reproduce research. 

We anticipate the nature of having longer papers reducing reproduction times is similar to that of the number of references included in the paper (\autoref{fig:xgboost_references}), where a strong positive relationship exists but instead plateaus after $\geq4$ references per page. In the case of References, having too few introduces an increase in reproduction time. Beyond providing empirical evidence for good scholastic behavior, references also serve as valuable sources of information for the techniques being built upon, allowing the reproducer to read additional papers that may further benefit their understanding of the paper they are attempting to reproduce.

It was previously noted that Tables had a significant relationship with the ability to reproduce a paper while Graphs did not. The Cox model does not contradict this conclusion from \citet{Raff2019_quantify_repro}, but it does refine our understanding. Having no, or less than one graph per page, corresponds to a lower reproduction time, and $\geq 1$ Graphs per page results in a minor increase in time. This appears to indicate that Graphs, while visually pleasing, may not be as effective of a visualization tool as the author might hope. We suspect that further delineating the nature and type of graphs would be valuable before drawing broader conclusions
. Tables, by comparison, predominantly help reproducibility. However, there is a narrow range in the number of tables per page that leads to a reduction in reproduction time, after which the value decays to become negligible. 

Of particular interest are the number of proofs (\autoref{fig:xgboost_proofs}) and equations (\autoref{fig:xgboost_equations}) due to their natural relationship with \textit{Theory} oriented papers. A new insight is that the more proofs in a paper the less time it takes to reproduce its results, but this value reaches a plateau near one proof for every two pages. This gives additional credence to arguments proposed by \citet{Sculley2018} on the importance of understanding the methods we develop at a scientific level. As can be seen in \autoref{fig:xgboost_proofs}, more proofs does naturally lend to more equations per page, which \autoref{fig:xgboost_equations} show has a non-linear relationship. Having no equations is bad, but having up to five per page is connected with a larger reduction in reproduction time than the benefit of more complete proofs. We note, however, that based on the protocol used it may be possible to obtain the benefits of both a moderate number of equations and well-proven works. The number of equations is counted only within the content of the main paper, but proofs are still counted when they appeared in appendices \cite{Raff2019_quantify_repro}. This suggests a potential ideal strategy for organizing the content of papers: few and judicious use of equations within the main paper to provide the necessary details for implementation and replication with more complete examination of the proofs in a separate section of the appendix. This we note should not be extrapolated to works that are purely theory based, which were beyond the scope of the original data. 

Last, we note some unexpected results under the Cox model compared to the original analysis. First, in \autoref{fig:xgboost_year} we can see that the amount of time to reproduce a paper has been increasing since 2005. This gives credence to the concern that we are entering a reproducibility crisis within machine learning, as argued by \citet{Hutson725} which \citet{Raff2019_quantify_repro} did not find evidence of.

\section{Conclusion} \label{sec:conclusion}

By extending a recent dataset with "reproduction times," measuring the number of days it took to reproduce a paper, we have proposed and evaluated the first survival analysis of machine learning research reproducibility. Combining this with more modern tree ensembles, we can use SHAP interpretability scores to gain new insights about what impacts the time it will take to reproduce a paper. In doing so, we obtain new insights into the utility of the most objective features of papers while also deriving new nuance to the nature of previously determined factors. We hope this will encourage others in this space to begin recording time spent on reproduction efforts as valuable information. 

\section*{Ethical Impact Statement}

Our work as the 
potential to positively influence the paper writing and publishing process, and in the near term, better equip us to study how reproducibility can be quantified. The timing of effort spent is intrinsically more objective than "success/non-success," as it does not preclude any failed attempt from eventually becoming successful. This is important at a foundational level to ensure machine learning operates with a scientific understanding of our results, and to understand the factors that may prevent others from successful replication. Indeed, by making the measure of reproducibility a more objective measure of time spent (though itself not perfect), we may increase the reproducibility of research studies. 

With these results come an important caveat that has been emphasized in the paper; that these results do not have exact timing information but rather a proxy measure, and we must use the survival model to circumvent missing information from the failed replication cases. Our analysis also can not explore all possible alternative hypothesis of the results. We risk readers taking these statements as a dogmatic truth and altering their research practices for the worse if it is not fully understood that this work provides the foundation for a new means of studying reproducibility. 

Our analysis must not be taken as a final statement on the nature of these relationships but instead as initial evidence. There are many confounding and unobserved factors that will impact the results and cannot be obtained. It is necessary that the reader understand these to be directions for future research with thought and care to design study protocols that can unravel these confounders. Further still, this study's data represents only attempts to obtain the same results purported in prior works. This is distinct from whether the conclusions from prior papers may be in error due to insufficient effort at tuning baselines or other factors. This is a distinct and important problem beyond the scope of this work, but has been found relevant to a number of sub-domains \cite{Dacrema2019,mlsys2020_73,Musgrave2020}. 

The caveats are amplified because the study is based on work with one reproducer. As such, it may not generalize to other individuals, and it may not generalize to other sub-disciplines within machine learning. To do so would require significant reproduction attempts from many individuals with varied backgrounds (e.g. training, years experience, education) and with careful controls. The original study was over a span of 8 years, so progress is unlikely unless a communal effort to collect this information occurs. A cohort of \textit{only 125 participants} would constitute \textit{a millennia of person effort} to make a better corpus over the same conditions. Our hope is that this work, by showing how we can better quantify reproduction attempts via time spent, gives the requisite objective needed to begin such larger communal efforts. 

\section*{Acknowledgements}

I'd like to thank Drew Farris, Frank Ferraro, Cynthia Matuszek, and Ashley Klein for their feedback and advice on drafts of this work. I would also like to thank Reviewer \#3 of my NeurIPS paper for spawning this follow-up work. I woefully underestimated how long it would take and the analyze "how long it took to reproduce". I hope this paper is a satisficing followup.

\bibliography{references}

\begin{thebibliography}{37}
\providecommand{\natexlab}[1]{#1}
\providecommand{\url}[1]{\texttt{#1}}
\providecommand{\urlprefix}{URL }
\expandafter\ifx\csname urlstyle\endcsname\relax
  \providecommand{\doi}[1]{doi:\discretionary{}{}{}#1}\else
  \providecommand{\doi}{doi:\discretionary{}{}{}\begingroup
  \urlstyle{rm}\Url}\fi

\bibitem[{Akiba et~al.(2019)Akiba, Sano, Yanase, Ohta, and
  Koyama}]{Akiba:2019:ONH:3292500.3330701}
Akiba, T.; Sano, S.; Yanase, T.; Ohta, T.; and Koyama, M. 2019.
\newblock {Optuna: A Next-generation Hyperparameter Optimization Framework}.
\newblock In \emph{Proceedings of the 25th ACM SIGKDD International Conference
  on Knowledge Discovery {\&} Data Mining}, KDD '19, 2623--2631. New York, NY,
  USA: ACM.
\newblock ISBN 978-1-4503-6201-6.
\newblock \doi{10.1145/3292500.3330701}.

\bibitem[{Bach et~al.(2015)Bach, Binder, Montavon, Klauschen, M{\"{u}}ller, and
  Samek}]{Bach2015}
Bach, S.; Binder, A.; Montavon, G.; Klauschen, F.; M{\"{u}}ller, K.-R.; and
  Samek, W. 2015.
\newblock {On Pixel-Wise Explanations for Non-Linear Classifier Decisions by
  Layer-Wise Relevance Propagation}.
\newblock \emph{PLOS ONE} 10(7).
\newblock ISSN 1932-6203.
\newblock \doi{10.1371/journal.pone.0130140}.

\bibitem[{Barba(2019)}]{Barba2019}
Barba, L.~A. 2019.
\newblock {Praxis of Reproducible Computational Science}.
\newblock \emph{Computing in Science {\&} Engineering} 21(1): 73--78.
\newblock ISSN 1521-9615.
\newblock \doi{10.1109/MCSE.2018.2881905}.

\bibitem[{Blalock et~al.(2020)Blalock, Gonzalez~Ortiz, Frankle, and
  Guttag}]{mlsys2020_73}
Blalock, D.; Gonzalez~Ortiz, J.~J.; Frankle, J.; and Guttag, J. 2020.
\newblock {What is the State of Neural Network Pruning?}
\newblock In \emph{Proceedings of Machine Learning and Systems 2020}, 129--146.

\bibitem[{Bou-Hamad, Larocque, and Ben-Ameur(2011)}]{Bou-Hamad2011}
Bou-Hamad, I.; Larocque, D.; and Ben-Ameur, H. 2011.
\newblock {A review of survival trees}.
\newblock \emph{Statist. Surv.} 5: 44--71.
\newblock ISSN 1935-7516.
\newblock \doi{10.1214/09-SS047}.

\bibitem[{Bouthillier, Laurent, and Vincent(2019)}]{pmlr-v97-bouthillier19a}
Bouthillier, X.; Laurent, C.; and Vincent, P. 2019.
\newblock {Unreproducible Research is Reproducible}.
\newblock In Chaudhuri, K.; and Salakhutdinov, R., eds., \emph{Proceedings of
  the 36th International Conference on Machine Learning}, volume~97 of
  \emph{Proceedings of Machine Learning Research}, 725--734. Long Beach,
  California, USA: PMLR,
  \urlprefix\url{http://proceedings.mlr.press/v97/bouthillier19a.html}.

\bibitem[{Collaboration(2015)}]{aac4716}
Collaboration, O.~S. 2015.
\newblock {Estimating the reproducibility of psychological science}.
\newblock \emph{Science} 349(6251).
\newblock ISSN 0036-8075.
\newblock \doi{10.1126/science.aac4716}.

\bibitem[{Dacrema, Cremonesi, and Jannach(2019)}]{Dacrema2019}
Dacrema, M.~F.; Cremonesi, P.; and Jannach, D. 2019.
\newblock {Are we really making much progress? A Worrying Analysis of Recent
  Neural Recommendation Approaches}.
\newblock In \emph{Proceedings of the 13th ACM Conference on Recommender
  Systems - RecSys '19}, 101--109. New York, New York, USA: ACM Press.
\newblock ISBN 9781450362436.
\newblock \doi{10.1145/3298689.3347058}.

\bibitem[{Datta, Sen, and Zick(2016)}]{Datta2016}
Datta, A.; Sen, S.; and Zick, Y. 2016.
\newblock {Algorithmic Transparency via Quantitative Input Influence: Theory
  and Experiments with Learning Systems}.
\newblock In \emph{2016 IEEE Symposium on Security and Privacy (SP)}, 598--617.
  IEEE.
\newblock ISBN 978-1-5090-0824-7.
\newblock \doi{10.1109/SP.2016.42}.

\bibitem[{Forde et~al.(2018)Forde, Head, Holdgraf, Panda, Perez, Nalvarte,
  Ragan-kelley, and Sundell}]{Forde2018}
Forde, J.; Head, T.; Holdgraf, C.; Panda, Y.; Perez, F.; Nalvarte, G.;
  Ragan-kelley, B.; and Sundell, E. 2018.
\newblock {Reproducible Research Environments with repo2docker}.
\newblock In \emph{Reproducibility in ML Workshop, ICML'18}.

\bibitem[{Gebru et~al.(2018)Gebru, Morgenstern, Vecchione, Vaughan, Wallach,
  Daume{\'{e}}, and Crawford}]{Gebru2018}
Gebru, T.; Morgenstern, J.; Vecchione, B.; Vaughan, J.~W.; Wallach, H.;
  Daume{\'{e}}, H.; and Crawford, K. 2018.
\newblock {Datasheets for Datasets}.
\newblock \emph{ArXiv e-prints} 1--27,
  \urlprefix\url{http://arxiv.org/abs/1803.09010}.

\bibitem[{Glenn and P.A.(2015)}]{Glenn2015}
Glenn, B.~C.; and P.A., I.~J. 2015.
\newblock {Reproducibility in Science: Improving the Standard for Basic and
  Preclinical Research}.
\newblock \emph{Circulation Research} 116(1): 116--126.
\newblock \doi{10.1161/CIRCRESAHA.114.303819}.

\bibitem[{Gronenschild et~al.(2012)Gronenschild, Habets, Jacobs, Mengelers,
  Rozendaal, van Os, and Marcelis}]{10.1371/journal.pone.0038234}
Gronenschild, E. H. B.~M.; Habets, P.; Jacobs, H. I.~L.; Mengelers, R.;
  Rozendaal, N.; van Os, J.; and Marcelis, M. 2012.
\newblock {The Effects of FreeSurfer Version, Workstation Type, and Macintosh
  Operating System Version on Anatomical Volume and Cortical Thickness
  Measurements}.
\newblock \emph{PLOS ONE} 7(6): 1--13.
\newblock \doi{10.1371/journal.pone.0038234}.

\bibitem[{Gundersen and Kjensmo(2018)}]{Gundersen2018}
Gundersen, O.~E.; and Kjensmo, S. 2018.
\newblock {State of the Art: Reproducibility in Artificial Intelligence}.
\newblock \emph{Proceedings of the 32nd AAAI Conference on Artificial
  Intelligence (AAAI-18)} 1644--1651.

\bibitem[{Harrell~Jr., Lee, and Mark(1996)}]{HarrellJr.1996}
Harrell~Jr., F.~E.; Lee, K.~L.; and Mark, D.~B. 1996.
\newblock {Multivariable Prognostic Models: Issues in Developing Models,
  Evaluating Assumptions and Adequacy, and Measuring and Reducing Errors}.
\newblock \emph{Statistics in Medicine} 15(4): 361--387.
\newblock ISSN 0277-6715.
\newblock
  \doi{10.1002/(SICI)1097-0258(19960229)15:4<361::AID-SIM168>3.0.CO;2-4}.

\bibitem[{Hutson(2018)}]{Hutson725}
Hutson, M. 2018.
\newblock {Artificial intelligence faces reproducibility crisis}.
\newblock \emph{Science} 359(6377): 725--726.
\newblock ISSN 0036-8075.
\newblock \doi{10.1126/science.359.6377.725}.

\bibitem[{Ioannidis(2017)}]{Ioannidis2017}
Ioannidis, J.~P. 2017.
\newblock {The Reproducibility Wars: Successful, Unsuccessful, Uninterpretable,
  Exact, Conceptual, Triangulated, Contested Replication}.
\newblock \emph{Clinical Chemistry} 63(5): 943--945.
\newblock ISSN 0009-9147.
\newblock \doi{10.1373/clinchem.2017.271965}.

\bibitem[{Ioannidis(2018)}]{Ioannidis2018}
Ioannidis, J. P.~A. 2018.
\newblock {Meta-research: Why research on research matters}.
\newblock \emph{PLOS Biology} 16(3): e2005468,
  \urlprefix\url{https://doi.org/10.1371/journal.pbio.2005468}.

\bibitem[{Kaplan and Meier(1958)}]{Kaplan1958}
Kaplan, E.~L.; and Meier, P. 1958.
\newblock {Nonparametric Estimation from Incomplete Observations}.
\newblock \emph{Journal of the American Statistical Association} 53(282):
  457--481.
\newblock ISSN 0162-1459.
\newblock \doi{10.1080/01621459.1958.10501452}.

\bibitem[{LeBlanc and Crowley(1992)}]{LeBlanc1992}
LeBlanc, M.; and Crowley, J. 1992.
\newblock {Relative Risk Trees for Censored Survival Data}.
\newblock \emph{Biometrics} 48(2): 411.
\newblock ISSN 0006341X.
\newblock \doi{10.2307/2532300}.

\bibitem[{Lin and Wei(1989)}]{Lin1989}
Lin, D.~Y.; and Wei, L.~J. 1989.
\newblock {The Robust Inference for the Cox Proportional Hazards Model}.
\newblock \emph{Journal of the American Statistical Association} 84(408):
  1074--1078.
\newblock ISSN 01621459.
\newblock \doi{10.2307/2290085}.

\bibitem[{Lipovetsky and Conklin(2001)}]{Lipovetsky2001}
Lipovetsky, S.; and Conklin, M. 2001.
\newblock {Analysis of regression in game theory approach}.
\newblock \emph{Applied Stochastic Models in Business and Industry} 17(4):
  319--330.
\newblock ISSN 1524-1904.
\newblock \doi{10.1002/asmb.446}.

\bibitem[{Lundberg, Erion, and Lee(2018)}]{Lundberg2018}
Lundberg, S.~M.; Erion, G.~G.; and Lee, S.-I. 2018.
\newblock {Consistent Individualized Feature Attribution for Tree Ensembles}.
\newblock \emph{ArXiv e-prints} (2),
  \urlprefix\url{http://arxiv.org/abs/1802.03888}.

\bibitem[{Lundberg and Lee(2017)}]{NIPS2017_7062}
Lundberg, S.~M.; and Lee, S.-I. 2017.
\newblock {A Unified Approach to Interpreting Model Predictions}.
\newblock In Guyon, I.; Luxburg, U.~V.; Bengio, S.; Wallach, H.; Fergus, R.;
  Vishwanathan, S.; and Garnett, R., eds., \emph{Advances in Neural Information
  Processing Systems 30}, 4765--4774. Curran Associates, Inc.
\newblock
  \urlprefix\url{http://papers.nips.cc/paper/7062-a-unified-approach-to-interpreting-model-predictions.pdf}.

\bibitem[{Mantel(1966)}]{PMID:5910392}
Mantel, N. 1966.
\newblock {Evaluation of survival data and two new rank order statistics
  arising in its consideration}.
\newblock \emph{Cancer chemotherapy reports} 50(3): 163—170.
\newblock ISSN 0069-0112,
  \urlprefix\url{http://europepmc.org/abstract/MED/5910392}.

\bibitem[{Mesnard and Barba(2017)}]{Mesnard2017}
Mesnard, O.; and Barba, L.~A. 2017.
\newblock {Reproducible and Replicable Computational Fluid Dynamics: It’s
  Harder Than You Think}.
\newblock \emph{Computing in Science {\&} Engineering} 19(4): 44--55.
\newblock ISSN 1521-9615.
\newblock \doi{10.1109/MCSE.2017.3151254}.

\bibitem[{Musgrave, Belongie, and Lim(2020)}]{Musgrave2020}
Musgrave, K.; Belongie, S.; and Lim, S.-N. 2020.
\newblock {A Metric Learning Reality Check}.
\newblock \emph{arXiv} \urlprefix\url{http://arxiv.org/abs/2003.08505}.

\bibitem[{Olorisade, Brereton, and Andras(2018)}]{Olorisade2018}
Olorisade, B.~K.; Brereton, P.; and Andras, P. 2018.
\newblock {Reproducibility in Machine Learning-Based Studies : An Example of
  Text Mining}.
\newblock In \emph{Reproducibility in ML Workshop, ICML'18}.

\bibitem[{Prinz, Schlange, and Asadullah(2011)}]{Prinz2011}
Prinz, F.; Schlange, T.; and Asadullah, K. 2011.
\newblock {Believe it or not: how much can we rely on published data on
  potential drug targets?}
\newblock \emph{Nature Reviews Drug Discovery} 10(9): 712.
\newblock ISSN 1474-1784.
\newblock \doi{10.1038/nrd3439-c1}.

\bibitem[{Raff(2017)}]{JMLR:v18:16-131}
Raff, E. 2017.
\newblock {JSAT: Java Statistical Analysis Tool, a Library for Machine
  Learning}.
\newblock \emph{Journal of Machine Learning Research} 18(23): 1--5,
  \urlprefix\url{http://jmlr.org/papers/v18/16-131.html}.

\bibitem[{Raff(2019)}]{Raff2019_quantify_repro}
Raff, E. 2019.
\newblock {A Step Toward Quantifying Independently Reproducible Machine
  Learning Research}.
\newblock In \emph{NeurIPS}. \urlprefix\url{http://arxiv.org/abs/1909.06674}.

\bibitem[{Ribeiro, Singh, and
  Guestrin(2016)}]{Ribeiro:2016:WIT:2939672.2939778}
Ribeiro, M.~T.; Singh, S.; and Guestrin, C. 2016.
\newblock {"Why Should I Trust You?": Explaining the Predictions of Any
  Classifier}.
\newblock In \emph{Proceedings of the 22nd ACM SIGKDD International Conference
  on Knowledge Discovery and Data Mining}, KDD '16, 1135--1144. New York, NY,
  USA: ACM.
\newblock ISBN 978-1-4503-4232-2.
\newblock \doi{10.1145/2939672.2939778}.

\bibitem[{Sculley et~al.(2015)Sculley, Holt, Golovin, Davydov, Phillips, Ebner,
  Chaudhary, Young, Crespo, and Dennison}]{10.5555/2969442.2969519}
Sculley, D.; Holt, G.; Golovin, D.; Davydov, E.; Phillips, T.; Ebner, D.;
  Chaudhary, V.; Young, M.; Crespo, J.-F.; and Dennison, D. 2015.
\newblock {Hidden Technical Debt in Machine Learning Systems}.
\newblock In \emph{Proceedings of the 28th International Conference on Neural
  Information Processing Systems - Volume 2}, NIPS'15, 2503–2511. Cambridge,
  MA, USA: MIT Press.

\bibitem[{Sculley et~al.(2018)Sculley, Snoek, Rahimi, and
  Wiltschko}]{Sculley2018}
Sculley, D.; Snoek, J.; Rahimi, A.; and Wiltschko, A. 2018.
\newblock {Winner's Curse? On Pace, Progress, and Empirical Rigor}.
\newblock In \emph{ICLR Workshop track}.
\newblock ISBN 1559-713X,
  \urlprefix\url{https://openreview.net/pdf?id=rJWF0Fywf}.

\bibitem[{Shrikumar, Greenside, and Kundaje(2017)}]{Shrikumar2017}
Shrikumar, A.; Greenside, P.; and Kundaje, A. 2017.
\newblock {Learning important features through propagating activation
  differences}.
\newblock In \emph{34th International Conference on Machine Learning, ICML
  2017}, 4844--4866.
\newblock ISBN 9781510855144.

\bibitem[{{\v{S}}trumbelj and
  Kononenko(2014)}]{Strumbelj:2014:EPM:2687513.2687579}
{\v{S}}trumbelj, E.; and Kononenko, I. 2014.
\newblock {Explaining Prediction Models and Individual Predictions with Feature
  Contributions}.
\newblock \emph{Knowl. Inf. Syst.} 41(3): 647--665.
\newblock ISSN 0219-1377.
\newblock \doi{10.1007/s10115-013-0679-x}.

\bibitem[{Wicherts et~al.(2006)Wicherts, Borsboom, Kats, and
  Molenaar}]{Wicherts2006}
Wicherts, J.~M.; Borsboom, D.; Kats, J.; and Molenaar, D. 2006.
\newblock {The poor availability of psychological research data for
  reanalysis.}
\newblock \emph{American Psychologist} 61(7): 726--728.
\newblock ISSN 1935-990X.
\newblock \doi{10.1037/0003-066X.61.7.726}.

\end{thebibliography}

\clearpage

\begin{appendix}

\section{Non-linear Error in Linear Cox Model}

Below is the plots showing the residuals for the two features that failed the cox model test. In both cases the errors appear to be mildly non-linear, causing the test to fail, but still show that the cox model was a reasonable fit to the data. 

\begin{figure}[!htbp]
\centering
\setlength{\lineskip}{\medskipamount}
\subcaptionbox{Normalized Number of Equations.\label{fig:cox_eq_residual_fit}}
    {
        \adjustbox{max width=\columnwidth}{%
	    \input{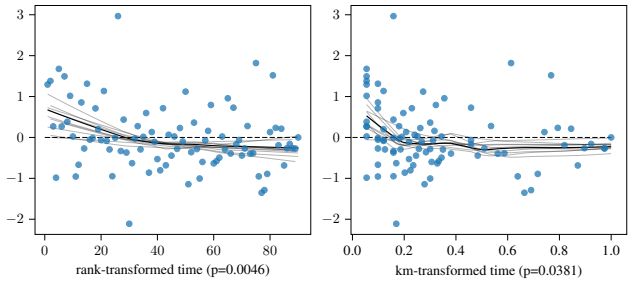}
	    }
    }
\subcaptionbox{Year First Attempted.\label{fig:cox_year_residual_fit}}
    {
        \adjustbox{max width=\columnwidth}{%
	    \input{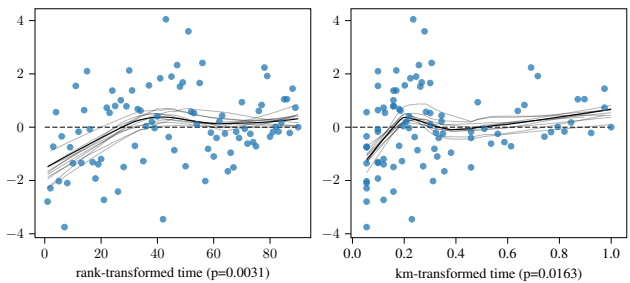}
	    }
    }
\caption{Residual errors for for the two variables which differed from the null hypothesis.
} \label{fig:cox_residual_fits}
\end{figure}

\section{XGBoost Code Snippet}

Here we include snippets of code used to produce the results of this paper so that their exact details are available and can be easily re-implemented. First, the below code specifies the functions used for training the XGBoost model and selecting the parameters via cross validation. The Optuna library was used to perform this parameter search. The function c\_statistic\_harrell implements the Concordance score referenced in \autoref{sec:linear_cox}, and the c\_eval function wraps the concordance calculation in a manner that XGBoost will accept as a target metric. 

\begin{minted}[
frame=lines,
framesep=2mm,
baselinestretch=1.2,
%
fontsize=\footnotesize,
tabsize=1, 
breaklines,
%
]{python}
def c_statistic_harrell(pred, labels):
  total = 0
  matches = 0
  for i in range(len(labels)):
    for j in range(len(labels)):
      if labels[j] > 0 and abs(labels[i]) > labels[j]:
        total += 1
        if pred[j] > pred[i]:
          matches += 1
  return matches/total

def c_eval(predt: np.ndarray, dtrain: xgboost.DMatrix):
  y = dtrain.get_label()
  return 'c', c_statistic_harrell(predt, y)

def objective(t):
  """
  t: the optuna trial object
  """
  depth = t.suggest_int('max_depth', 3, 10)
  eta = t.suggest_loguniform('eta', 0.003, 0.5)
  subsample = t.suggest_uniform("subsample", 0.2, 0.7)
  rounds = t.suggest_int('rounds', 10, 300)
  colsample_bytree = t.suggest_uniform("colsample_bytree", 0.3, 1.0)
  colsample_bylevel = t.suggest_uniform("colsample_bylevel", 0.5, 1.0)
  lambda_ = t.suggest_uniform("lambda", 0.1, 2.0)

  params = {
    "eta": eta,
    "max_depth": depth, 
    "objective": "survival:cox",
    "subsample": subsample,
    "colsample_bytree": colsample_bytree,
    "colsample_bylevel": colsample_bylevel,
    "lambda": lambda_,
    'silent':1
    }

  cv_results = xgboost.cv(
    params,
    xgb_full_ord,
    num_boost_round=rounds,
    seed=42,
    nfold=10,
    feval=c_eval
  )
    
  return cv_results.tail(n=1)['test-c-mean'].values[0]

study = optuna.create_study(direction='maximize')
study.optimize(objective, n_trials=100)
\end{minted}

After running this search, the below parameters are the result found when we ran the above code and are used for all figures. 

\begin{minted}[
frame=lines,
framesep=2mm,
baselinestretch=1.2,
%
fontsize=\footnotesize,
tabsize=1, 
breaklines,
%
]{python}
params = {
  "objective": "survival:cox",
  "eta": 0.02126844892731846,
  "max_depth": 8, 
  "subsample": 0.20210001379297854,
  'colsample_bytree': 0.37002203782589316, 
  'colsample_bylevel': 0.7085528974300124,  
  'lambda': 1.497998138207469,
  'silent': 1
}
boost_rounds = 120
\end{minted}

\section{Replication of Cox Analysis using Median instead of Mean} \label{sec:by_median}

We have stated that using the mean vs median provides different but qualitatively similar results. In this section, we show the results comparing the imputation by mean value (done in the main paper) versus the median reproduction time (here in this appendix). For all results, the mean will be shown on the left and median on the right. Starting with \autoref{tbl:p_val_median}, we can see that conclusions on significance are not impacted at our threshold $\alpha$. 

\begin{table}[!h]
\caption{Comparing p-values  when training a linear Cox proportional hazard model for each feature's importance between the Cox model with non-reproductions estimated as $\geq$ mean (middle column) and $\geq$ median (right column).
}
\label{tbl:p_val_median}
\centering
\adjustbox{width=\columnwidth}{%
\begin{tabular}{lrr}
\hline
\multicolumn{1}{c}{Feature} & \multicolumn{1}{c}{Cox by Mean} & \multicolumn{1}{c}{Cox by Median} \\ \hline
Year Published              & 0.92                            & 0.62                              \\
Year Attempted              & 0.45                            & 0.46                              \\
Has Appendix                & \textbf{0.07}                   & \textbf{0.05}                     \\
Uses Exemplar Toy Problem   & 0.20                            & 0.28                              \\
Looks Intimidating          & 0.20                            & 0.25                              \\
Exact Compute Used          & 0.39                            & 0.49                              \\
Data Available              & 0.81                            & 0.98                              \\
Code Available              & 0.18                            & 0.28                              \\
Number of Authors           & 0.68                            & 0.47                              \\
Pages                       & 0.82                            & 0.79                              \\
Num References              & 0.54                            & 0.47                              \\
Number of Equations         & 0.74                            & 0.80                              \\
Number of Proofs            & 0.47                            & 0.55                              \\
Number of Tables            & 0.86                            & 0.89                              \\
Number of Graphs/Plots      & 0.51                            & 0.63                              \\
Number of Other Figures     & 0.98                            & 0.85                              \\
Conceptualization Figures   & 0.13                            & 0.17                              \\
Hyperparameters Specified   & 0.49                            & 0.34                              \\
Algorithm Difficulty        & \textbf{0.03}                   & \textbf{0.07}                     \\
Paper Readability           & \textbf{\textless{}0.005}       & \textbf{\textless{}0.005}         \\
Pseudo Code                 & \textbf{\textless{}0.005}       & \textbf{\textless{}0.005}         \\
Compute Needed              & 1.000                           & 0.98                              \\
Rigor vs Empirical          & \textbf{0.02}                   & \textbf{0.02}                     \\ \hline
\end{tabular}
}
\end{table}

In \autoref{fig:xgboost_cox_median_cmpr1} and \autoref{fig:xgboost_cox_median_cmpr2}, we show the SHAP scores for each feature in detail from an XGboost tree optimized by Optuna. The Median based results had a slightly lower cross-validated concordance of 0.76 but are otherwise similar in quality. 

\begin{figure*}[!htb]
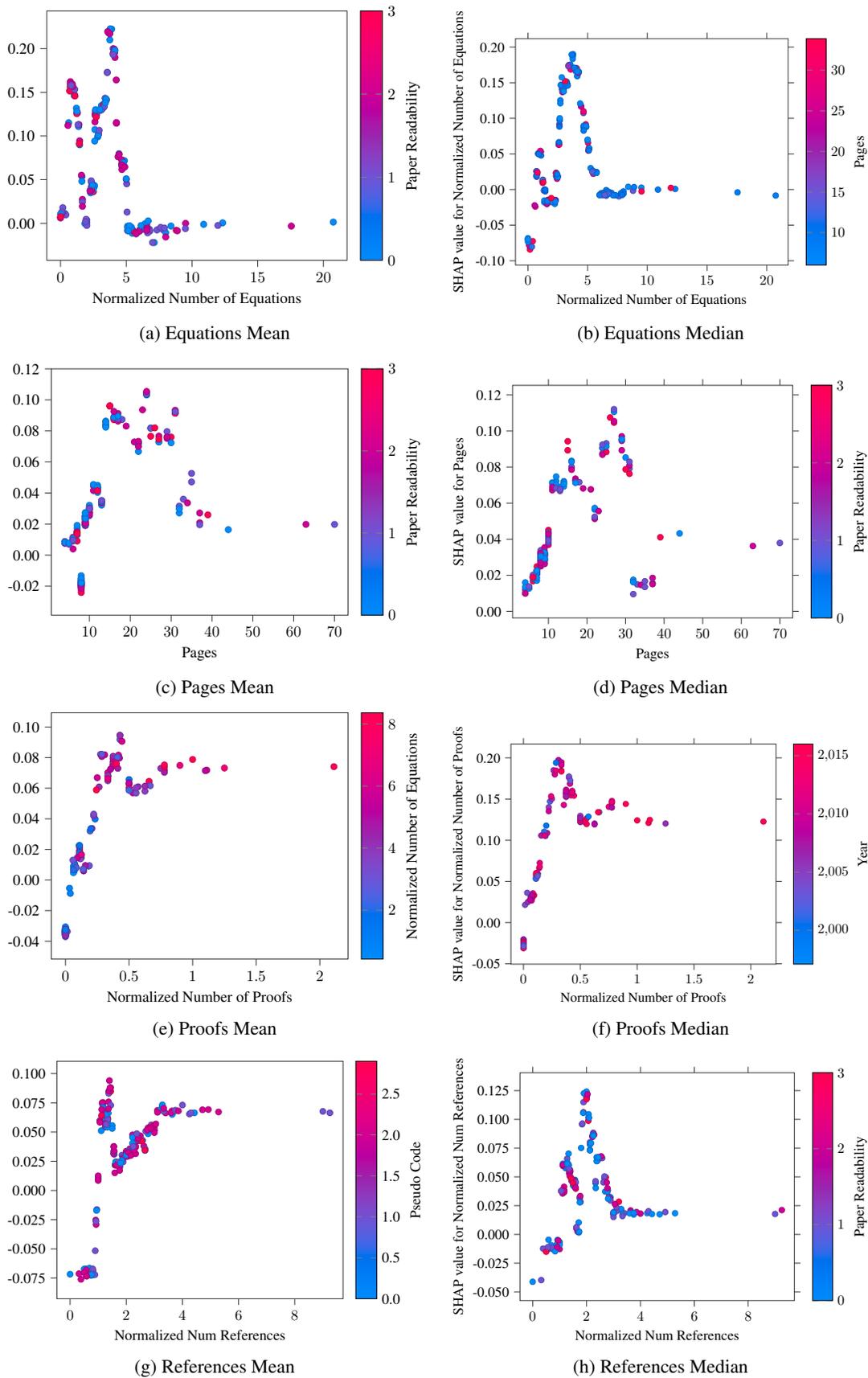

\centering
\setlength{\lineskip}{\medskipamount}
\subcaptionbox{Equations Mean}
    {
        \adjustbox{width=0.40\textwidth}{%
	    \input{figs/shap/xgboost_shape_decor_Normalized_Number_of_Equations.tex}
	    }
    }
\subcaptionbox{Equations Median}
    {
        \adjustbox{width=0.40\textwidth}{%
	    \input{figs_med/median_xgboost_shape_decor_Normalized_Number_of_Equations.tex}
	    }
    }
\subcaptionbox{Pages Mean}
    {
        \adjustbox{max width=0.40\textwidth}{%
	    \input{figs/shap/xgboost_shape_decor_Pages.tex}
	    }
    }
\subcaptionbox{Pages Median}
    {
        \adjustbox{max width=0.40\textwidth}{%
	    \input{figs_med/median_xgboost_shape_decor_Pages.tex}
	    }
    }
\subcaptionbox{Proofs Mean}
    {
      \adjustbox{max width=0.40\textwidth}{%
	    \input{figs/shap/xgboost_shape_decor_Normalized_Number_of_Proofs.tex}
	    }
    }
\subcaptionbox{Proofs Median}
    {
      \adjustbox{max width=0.40\textwidth}{%
	    \input{figs_med/median_xgboost_shape_decor_Normalized_Number_of_Proofs.tex}
	    }
    }
\subcaptionbox{References Mean}
    {
      \adjustbox{max width=0.40\textwidth}{%
	    \input{figs/shap/xgboost_shape_decor_Normalized_Num_References.tex}
	    }
    }
\subcaptionbox{References Median}
    {
      \adjustbox{max width=0.40\textwidth}{%
	    \input{figs_med/median_xgboost_shape_decor_Normalized_Num_References.tex}
	    }
    }
\caption{SHAP individual features (change in log-hazard ratio) for several numeric features on the y-axis with feature values on the x-axis. Each figure has a color set based on the value of a second feature, indicated on the right. The second feature is selected by having the highest SHAP interaction. The left column shows the results when using the mean to impute a right censored value for non-reproduced papers. The right column shows the results when using the median. 
} \label{fig:xgboost_cox_median_cmpr1}
\end{figure*}

\begin{figure*}[!htb]
\centering
\setlength{\lineskip}{\medskipamount}
\subcaptionbox{Tables Mean}
    {
      \adjustbox{max width=0.2301\textwidth}{%
	    \input{figs/shap/xgboost_shape_decor_Normalized_Number_of_Tables.tex}
	    }
    }
\subcaptionbox{Tables Median}
    {
      \adjustbox{max width=0.2301\textwidth}{%
	    \input{figs_med/median_xgboost_shape_decor_Normalized_Number_of_Tables.tex}
	    }
    }
\subcaptionbox{Graphs Mean}
    {
      \adjustbox{max width=0.2301\textwidth}{%
	    \input{figs/shap/xgboost_shape_decor_Normalized_Number_of_Graphs_Plots.tex}
	    }
    }
\subcaptionbox{Graphs Median}
    {
      \adjustbox{max width=0.2301\textwidth}{%
	    \input{figs_med/median_xgboost_shape_decor_Normalized_Number_of_Graphs_Plots.tex}
	    }
    }
\subcaptionbox{Year of publication Mean}
    {
      \adjustbox{max width=0.2301\textwidth}{%
	    \input{figs/shap/xgboost_shape_decor_Year.tex}
	    }
    }
\subcaptionbox{Year of publication Median}
    {
      \adjustbox{max width=0.2301\textwidth}{%
	    \input{figs_med/median_xgboost_shape_decor_Year.tex}
	    }
    }
\subcaptionbox{Year Attempted Mean}
    {
      \adjustbox{max width=0.2301\textwidth}{%
	    \input{figs/shap/xgboost_shape_decor_Year_Attempted.tex}
	    }
    }
\subcaptionbox{Year Attempted Median}
    {
      \adjustbox{max width=0.2301\textwidth}{%
	    \input{figs_med/median_xgboost_shape_decor_Year_Attempted.tex}
	    }
    }
\subcaptionbox{Conceptualization Figures Mean}
    {
        \adjustbox{max width=0.2301\textwidth}{%
	    \input{figs/shap/xgboost_shape_decor_Normalized_Conceptualization_Figures.tex}
	    }
    }
\subcaptionbox{Conceptualization Figures Median}
    {
        \adjustbox{max width=0.2301\textwidth}{%
	    \input{figs_med/median_xgboost_shape_decor_Normalized_Conceptualization_Figures.tex}
	    }
    }

\caption{
Continuation of \autoref{fig:xgboost_cox_median_cmpr1}. 
} \label{fig:xgboost_cox_median_cmpr2}
\end{figure*}

\end{appendix}

\end{document}